\documentclass[10pt,twocolumn,letterpaper]{article}

\usepackage[pagenumbers]{wacv} % To force page numbers, e.g. for an arXiv version

\usepackage{pifont}% http://ctan.org/pkg/pifont
\newcommand{\cmark}{\ding{51}}%
\newcommand\xmark{\ding{55}}%
\usepackage{float}
\usepackage{soul}
\usepackage{wrapfig}
\usepackage{graphicx}
\usepackage{amsmath}
\usepackage{amssymb}
\usepackage{booktabs}

\usepackage[pagebackref,breaklinks,colorlinks]{hyperref}

\usepackage[capitalize]{cleveref}
\crefname{section}{Sec.}{Secs.}
\Crefname{section}{Section}{Sections}
\Crefname{table}{Table}{Tables}
\crefname{table}{Tab.}{Tabs.}

\def\wacvPaperID{2147} % *** Enter the WACV Paper ID here
\def\confName{WACV}
\def\confYear{2025}

\begin{document}

\title{DiHuR: Diffusion-Guided Generalizable Human Reconstruction}
\author{
% Jinnan Chen \qquad Chen Li \qquad Gim Hee Lee  \\
% Department of Computer Science, National University of Singapore\\
% {\tt\small \{jinnan.c, lichen\}@u.nus.edu \qquad gimhee.lee@nus.edu.sg}
% }
Jinnan Chen$^{1}$ \qquad Chen Li$^{2,3}$\thanks{Chen Li was at the National University of Singapore when this work was done.} \qquad Gim Hee Lee$^{1}$ \vspace{1mm} \\
 Department of Computer Science, National University of Singapore$^{1}$ \\ IHPC, Agency for Science, Technology and Research, Singapore$^{2}$ \\
 CFAR, Agency for Science, Technology and Research, Singapore$^{3}$ 
 \vspace{1mm} 
 \\
 \texttt{jinnan.c@u.nus.edu} \quad \texttt{lichen@u.nus.edu} \quad \texttt{gimhee.lee@comp.nus.edu.sg}
 }
 
% \author{Jinnan Chen\\
% \and
% Chen Li\\

\maketitle

%%%%%%%%% ABSTRACT
\begin{abstract}
We introduce DiHuR, a novel Diffusion-guided model for generalizable Human 3D Reconstruction and view synthesis from sparse, minimally overlapping images. While existing generalizable human radiance fields excel at novel view synthesis, they often struggle with comprehensive 3D reconstruction. Similarly, directly optimizing implicit Signed Distance Function (SDF) fields from sparse-view images typically yields poor results due to limited overlap. To enhance 3D reconstruction quality, we propose using learnable tokens associated with SMPL vertices to aggregate sparse view features and then to guide SDF prediction. These tokens learn a generalizable prior across different identities in training datasets, leveraging the consistent projection of SMPL vertices onto similar semantic areas across various human identities. This consistency enables effective knowledge transfer to unseen identities during inference. Recognizing SMPL's limitations in capturing clothing details, we incorporate a diffusion model as an additional prior to fill in missing information, particularly for complex clothing geometries. Our method integrates two key priors in a coherent manner: the prior from generalizable feed-forward models and the 2D diffusion prior, and it requires only multi-view image training, without 3D supervision. DiHuR demonstrates superior performance in both within-dataset and cross-dataset generalization settings, as validated on THuman, ZJU-MoCap, and HuMMan datasets compared to existing methods.
% DiHuR outperforms existing methods in generating high-fidelity 3D human reconstructions and novel view syntheses from challenging sparse-view inputs.
\end{abstract}
\section{Introduction}
\label{sec:intro}

Neural scene representation\cite{nerf20} enables the realistic generation of 3D digital models from 2D observations of the scene that are useful for many real-world applications such as Augmented and Virtual Reality (AR/VR), and the creation of digital avatars. %for building the metaverse. Among this, 3D reconstruction from 2D observations is the foundation. 
\begin{figure}[t] 
\centering 
\includegraphics[width=0.48\textwidth]{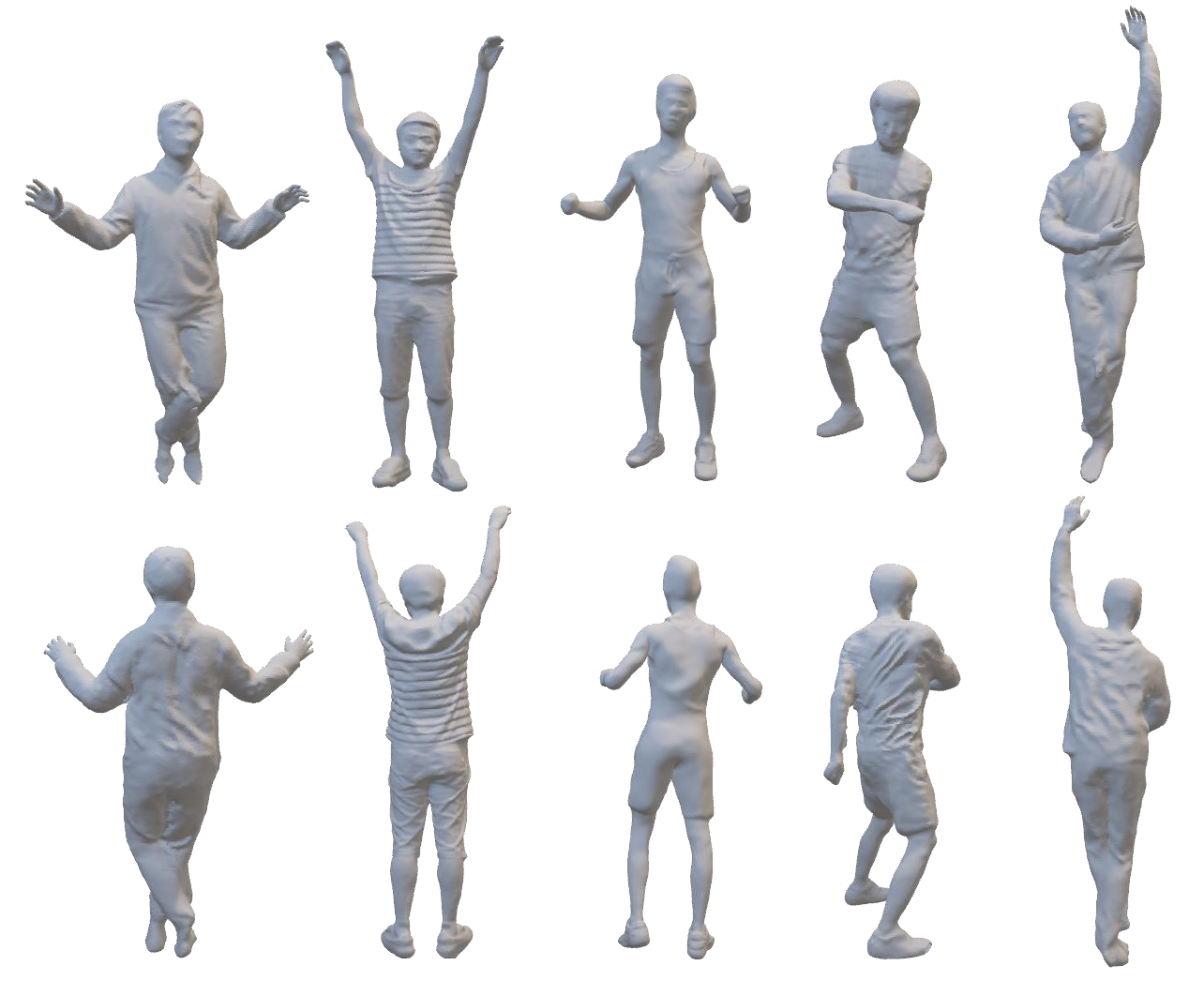} 
\vspace{-1mm}
\caption{Given 3 views of images with minimal overlap FOV, our method can accurately reconstruct 3D human models.}
\vspace{-1mm}
\label{fig:teaser}
\end{figure}
%
% Combining generalizable NeRF variants such as IBRNet~\cite{ibrnet21} with NeuS, SparseNeuS~\cite{sparseneus22} is proposed for fast and generalizable 3D reconstruction. %Although it makes reconstruction from sparse views possible, overlapping views are still a requirement, which means their target is the local surface. 
% Despite enabling neural reconstruction from sparse views, SparseNeuS still requires input images with large overlapping views. 
Although existing {generalizable human NeRF} \cite{nb21,anerf21,gpnerf22,mps22,arah22,sherf23} have achieved impressive, they mainly focus on novel view synthesis. The surfaces extracted from these fields are often unsatisfactory, particularly in sparse view settings. This is due to the little or non-existent overlap between views, which complicates feature aggregation. In such scenarios, it becomes challenging to identify the correct features with low variance(high confidence) across views. Consequently, the algorithm often resorts to simple averaging, leading to suboptimal results in surface reconstruction.
% Some work \cite{neus21,arah22} have shown effective 3D reconstruction, which however requires expensive {per-scene optimzation} from scratch. 
% In this work, we tackle the task of generalizable human reconstruction and novel view synthesis, where we aim to reconstruct detailed human surface from only sparse views.
In this paper, we focus on the challenging task of \textit{generalizable} 3D human reconstruction and NVS with images from sparse cameras. To this end, {we propose the %DiHuR 
\underline{\textbf{Di}}ffusion-guided Generalizable \underline{\textbf{Hu}}man \underline{\textbf{NeRF}} (DiHuR) framework}, which leverages the SMPL model~\cite{smpl15} to build geometric-consistent features to infer the 3D structure and incorporates pre-trained diffusion models as geometric guidance to enhance reconstruction quality. Specifically, we fuse the feature  by cross-view attention with the learnable tokens attached on the SMPL vertices. The fused features are further refined with several self-attention layers among all the tokens for information exchange. In order to sample the sparse features attached on the fixed number of SMPL vertices, we compute K nearest neighbours and average K features based on the distances. Although effective, the SMPL model is learned from minimal clothes body data and thus fails to model geometric details on the clothes. To solve this, we render the normal map from several target views and feed them to the pre-trained 2D diffusion model to compute SDS loss for fast finetuning. 
% We back-propagate this SDS loss to our generalizable model.
We adapt the diffusion model \cite{diffusion22} for super-resolution to provide detailed 3D geometry prior. We first render the normal map from our model, which is then upsampled 4x and used as the input of the diffusion model. The rendered map is also used together with the text as the condition for the denoising process. SDS loss is back-propagated to update our model parameters.
% Specifically, the geometry-aware code is used to infer view visibility that improves 3D geometry prediction by ensuring self-occlusion awareness in the predicted color blending weights.
%that are common in our sparse minimal overlapping FOV setting. 
% For color blending on the rays, we convert the SDF value to a globally consistent ray blending weight following \cite{neus21} since nearby points in the normal direction of a point also contribute to the conversion.
% 3D Human reconstruction from a single image is a well-studied~\cite{pifuhd20,pifu19,geopifu20,ipifu22,spifu22}. However, these methods require full supervision from large amounts of 3D models with corresponding 2D images to learn priors for the completion of occluded areas. In view of the difficulty in getting the 2D image-to-3D models training data, 
A multi-target optimization strategy is also proposed to implicitly regularize the 3D surface. We simultaneously sample rays from different camera views, focusing on the same body part. This ensures intersecting rays between views, which implicitly enforce multi-view consistency during surface prediction. We also apply a second-order Signed Distance Function (SDF) regularization to enhance surface smoothness. Fig~\ref{fig:teaser} shows examples of our accurate 3D reconstruction results on the sparse camera setting. 
% We can see that the 3D models from NeuS, SparseNeuS, and GP-NeRF are corrupted with artifacts while our DiHuR shows a result that is comparable to NeuS with dense views. 
Our main contributions can be summarized as follows: %\vspace{-1mm}
\begin{itemize}
    \item We propose DiHuR to solve the challenging task of generalizable 3D human reconstruction and NVS in the sparse views. %\vspace{-1mm}
    \item We propose to use learnable latent codes attached on SMPL vertices to guide the sparse view reconstruction and 2D prior from the diffusion model to enhance the details. %\vspace{-1mm}
     %using the second-order gradient of SDF to improve the surface reconstruction. %We show that better 3D regularization improves NVS quality by alleviating the 2D-3D ambiguity. %\vspace{-1mm}
    \item We achieve SOTA-performance for  3D human reconstruction, and novel view synthesis on the commonly used dataset in the sparse view settings.
    
\end{itemize}

\begin{figure*}[t] 
\centering 
\includegraphics[width=1\textwidth]{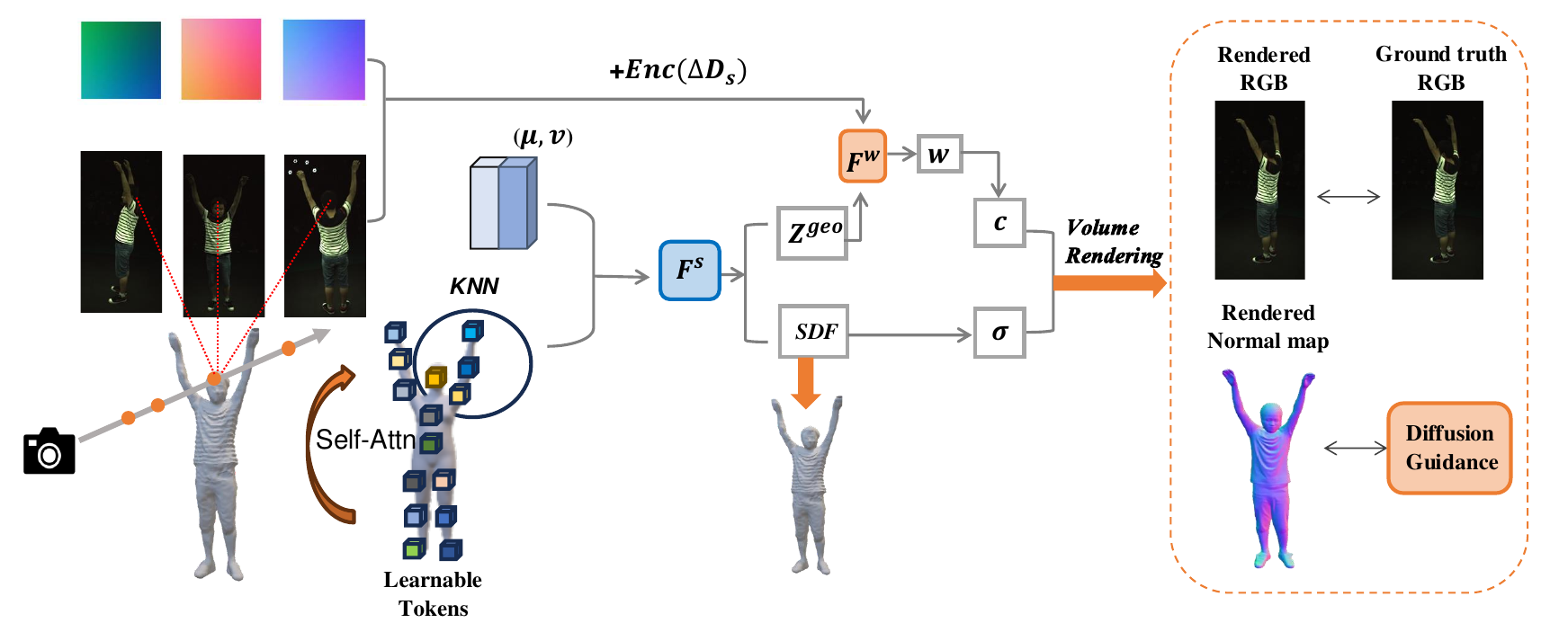} 
\vspace{-1mm}
\caption{The overview of our proposed DiHuR. Our pipeline mainly consists of three parts: 1) Feature aggregation, 2) SDF prediction, and 3) Appearance prediction. Our learnable tokens serve as query tokens to aggregate sparse view features, which is then interpolated by KNN for each query point. Volume rendering is done with aggregated features and mean, variance of the projected features from sparse input views. During inference, we finetune our model with normal SDS loss to enhance the details.}
\vspace{-1mm}
\label{fig:framework}
\end{figure*}

\section{Related works}
% \label{sec:formatting}
% ng (we discourage it), use 10-point Times, boldface, initially capitalized, flush left, preceded by one blank line, followed by a period, and your text on the same line.

\paragraph{Human NeRF and Gaussian.}
NB~\cite{nb21} first combines NeRF with a parametric human body model SMPL \cite{smpl15} to regularize the optimization process. Although it achieves surprising NVS results, it needs long-term optimization and is hard to generalize to unseen identities. Ani-NeRF \cite{animatable21} learns a canonical NeRF model and a backward LBS network which predicts residuals to the deterministic SMPL-based backward LBS (Linear Blending Skinning) to animate the learned human NeRF model. A-NeRF \cite{anerf21} designs a backward mapping with bone-relative representation for feature encoding, which also helps the initial pose correction. Subsequently, generalizable NeRFs are \cite{ibrnet21,mvsnerf21,pixelnerf20,sherf23} are proposed to make NeRF generalizable to novel scenes in a feed-forward way. They commonly condition the NeRF on the pixel-aligned image features and learn blending weights or directly infer the RGB color condition on such features. However, due to the non-rigid structure and complicated poses of the human body, directly applying such methods fails to work well on real human datasets. \cite{nhp21} utilize view transformers and temporal transformers to aggregate multi-view and multi-frame features to improve the model generalizability. \cite{gpnerf22} combine the SMPL model and generalizable NeRF in an efficient fashion. However, the 3D reconstruction is far from satisfying due to the lack of 3D constraints, and the NVS is blurry without awareness of the 3D geometry. More recently, people combine 3D gaussian splatting with human template model to reconstruct high quality humans \cite{chen2024generalizablehumangaussianssingleview,hu2023gauhumanarticulatedgaussiansplatting,liu2023humangaussian}. However, accurate surface extraction from noisy 3DGS remains a chanllenge. 
\paragraph{Human Surface Reconstruction.}
PIFu\cite{pifu19} is first proposed to reconstruct 3D model from single image with high quality.  Subsequently, PIFuHD\cite{pifuhd20} and ECON\cite{econ23} is proposed to improve geometric quality with multi-level pixel-aligned features and normal features as additional guidance. More recently, SiTH \cite{sith24} proposes to combine a back-view hallucination model with an SDF-based mesh reconstruction model. Similarly, SiFU\cite{sifu24} employs a text-to-image diffusion-based prior to generate realistic and consistent textures for invisible views.  However, these methods focus on single image reconstruction. Although these methods attempt to hallucinate reasonable interpretations of these invisible areas, the generated regions may not always align with users’ expectations. Also, ground truth 3D supervision are used for all of these methods. Compared to these methods, our approach can be trained solely on multi-view images and achieves better performance on sparse view settings.\\
\paragraph{Diffusion models as guidance.}
Recently, diffusion models \cite{diffusion22} have dominated the image generation area for its exceptional generative quality.
{Some works \cite{dream23,magic23,tech24} therefore utilize diffusion models as guidance to optimize an underlying} radiance field for 3D generation and achieve SOTA performance. Specifically, they introduce a Score Distillation Sampling (SDS) loss function, which formulates the same loss function used in the original diffusion model training process, but back-propagate the gradients towards the input rendered images and keeps the pre-trained diffusion models as fixed. With only this SDS loss as supervision, they can generate fantastic 3D meshes with text as an input condition. However, they often require lengthy optimization due to SDS loss computation and per-scene NeRF optimization from scratch. Inspired by these developments, our approach combines the SDS loss with an initial reconstruction from a generalizable NeRF model. This strategy preserves the prior information from pre-trained diffusion models and leverages the efficiency of feed-forward NeRF models, reducing overall optimization time.

\section{Our Method: DiHuR}
\paragraph{Problem Analysis.}
We propose a generalizable reconstruction approach for 3D human reconstruction with sparse cameras. The key %step
challenge in this generalizable setting is how to fuse the features extracted from multi-view images with minimal overlaps, \eg 3 cameras placed $120^\circ$ apart around the subject. A naive aggregation of the multi-view image features leads to high variance due to inconsistent features from self-occlusions. As a result, the network cannot differentiate between the self-occluded points and points in the free space. Without any supervision on the blending weights, the network can assign colors of occluded image pixels with high blending weights leading to inaccurate color prediction. To circumvent this problem, we propose to set learnable tokens anchored on the SMPL vertices which can learn the common prior across the training dataset to guide the SDF prediction. After the direct inference, we render the normal map from multi-views and finetune the partial parameters with normal SDS loss from the 2D pre-trained diffusion model as geometric guidance to enhance the details.
\vspace{-1mm}
\paragraph{Overview.}
As illustrated in Fig.~\ref{fig:framework}, our DiHuR is a volume rendering pipeline with SDF-based volume density representation. We first compute features for each sample point from two sources, \ie directly 3D-to-2D projected features' mean and variance, as well as sparse features aggregated from learnable tokens. These two features are concatenated and then used to predict SDF value along the ray and a geometry code that represents geometry embedding. For color prediction, use the geometry code and relative ray direction from each of the source camera centers for blending weights prediction. We use ground truth RGB images as supervision. During training, we use multiple target images simultaneously as the supervision to better regularize the underlying 3D geometry. During inference, we finetune our partial model parameters with normal SDS loss.

\subsection{Learnable Tokens as Human Prior}
Given $S$ source images $\{ I_{s} \}_{s=1}^{S}$ of a human captured by $S$ sparse cameras with minimal overlap FOV, we first use a multi-resolution image feature extractor to get multi-view features %, which are denoted as 
$\{F_{s} \}_{s=1}^{S}$. We use Plucker ray embedding to densely encode the camera poses. The collection of the RGB value and ray embedding for each pixel are concatenated into a 9-channel feature map as the input \(I=\{c_{i},o_{i} \times d_{i},d_{i} \mid i=1,2,...,N\}\), where i is the pixel index. In order to aggregate the accurate features from sparse views and make the feature occlusion-aware, we set learnable tokens attached on each of the SMPL vertices as $\{T_{q} \mid q=1, 2, ..., 6890\}$. Subsequently, we aggregate the features from $S$ source views for each vertex $p_{q}$ with a multi-head cross-view attention module to obtain the SMPL aggregated feature $F_{q}^{a}$ as:
\begin{equation}
\centering
\begin{aligned}
F_{q}^{a}=\sum_{s=1}^{S} w_{s}^{q}F_{s}(p_{q})
\end{aligned}
\label{eq:smpl_sum}
\end{equation}
\begin{equation}
\centering
\begin{aligned}
w_{s}^{q} = SoftMax(\frac{(L_{q}(T_{q} )(L_{k}(F_{s}(p_{q}))^\top}{\sqrt{d}}). 
 % w_{s}^{q}= frac{ mlp(T_{q}) mlp(F_{s}(p_{q}).T}{t /sqrt}
\end{aligned}
\label{eq:smpl_cross}
\end{equation}
,where $L_{q}$ $L_{k}$ are MLP for query and key embedding, $d$ is the model dimensions. Then, we add self-attention layers to let all the tokens' feature attend each other. Our approach leverages learnable tokens associated with SMPL vertices. These tokens acquire a generalizable prior across diverse identities during training and exploit the consistent mapping of SMPL vertices to similar semantic regions on different human bodies, which serves as a human prior facilitating knowledge transfer to novel identities during inference.

\subsection{SDF Prediction with KNN Features}
The SMPL feature is sparse in the 3D space since it only contains $6,890$ vertices. When we sample each point on the ray, we first identify K-Nearest Neighbours from all the SMPL vertices and compute the aggregated features based on the distance.
\begin{equation}
\centering
\begin{aligned}
 Fus(p)= \sum_{k=1}^{K} w_{k}F_{k}^{a}
\end{aligned}
\label{eq:smpl}
\end{equation}
where $w_{k}$ is computed with an inverse distance function and softmax function. To infer the SDF  value for a 3D point $p$, we project the 3D point to the source image spaces and compute the mean and variance of the image features, \ie $\mu=\frac{1}{S}\sum_{s=1}^{S}F_{s}(p)$ and $v=$Var$(\{ F_{s}(p)\}_{s=1}^{S} )$. We denote the concatenation of the mean and variance as the global feature $F^{glo}(p)= [ \mu,v ]$. The global feature is then concatenated with the KNN fused feature at $p$. The concatenated feature is fed into the SDF  prediction network $\mathcal F^{s }$ to predict the SDF  value:
\begin{equation}
\centering
\begin{aligned} 
\{s(p),Z^{geo}\}=\mathcal F^{s }(\gamma(p), [F^{glo}(p) ,Fus(p)] ),
\end{aligned}
\label{eq:SDF }
\end{equation}
where $\gamma$ denotes the positional encoding, and $s(p)$ denotes the predicted SDF  value at $p$. The network also outputs a geometry code $Z^{geo}$ which can be seen as an implicit local 3D structure encoding. This geometry code %will be
is used for the color prediction in the next step.
\subsection{Appearance Prediction}
% \paragraph{Directional encoding.}
% Following \cite{ibrnet21}, we take the difference between the target and source views into account during color prediction. Intuitively, 
% % between $d$ and $d_{k}$ 
% the color at the target view is more similar to the color at a view with a smaller view direction difference and vice versa. Specifically, we compute the difference between two views $d$ and $d_{k}$ as $\Delta d_{k}=d-d_{k}$. We encode this vector to a high dimension as $Enc(\Delta d)$. We then add the view direction feature to the original image features to obtain a new image feature expressed as:

\paragraph{Color prediction.}
We first blend the color among sparse views and then integrate the blended color along the ray to predict the final pixel color. 
We first compute the difference between two views $d$ and $d_{s}$ as $\Delta d_{s}=d-d_{s}$. We encode this vector to a high dimension as $Enc(\Delta d)$. We then add the view direction feature to the original image features to obtain a new image feature expressed as:
\begin{equation}
\centering
\begin{aligned} 
F^{\oplus}_{s}=F_{s}+Enc(\Delta d).
\end{aligned}
\label{eq:raydiff}
\end{equation}
% Blending weights prediction is quite essential for our setting because the reliable color can only be from 1-2 of the 4 sparse views with minimal overlap FOV and self-occlusion. To better predict the blending weights, we use the previous output geometry code as guidance. 
% Specifically, we first use a small MLP to predict the view visibility as:
% \begin{equation}
% \centering
% \begin{aligned} 
% v_{k}=\mathcal F^{v}(F^{\oplus}_{k},Z^{geo})\\
% \end{aligned}
% \label{eq:visi}
% \end{equation}
Then, we concatenate image feature $F^{\oplus}_{s}$, the geometry code $Z^{geo}$ %and the view visibility $v_{k}$ together, 
and feed it into our blending weights prediction network $\mathcal F^{w}$ for view blending weights $w_{s}$ prediction:

\begin{equation}
\centering
\begin{aligned} 
w_{s}=\mathcal F^{w}(F^{\oplus}_{s},Z^{geo}).\\
\end{aligned}
\label{eq:blendweights}
\end{equation}
This information contains rich geometry clues to help the network infer the correct blending weight for the source images with few color correspondences due to view sparsity. The color for each 3D point is the weighted sum of the projected color $c_{s}= I_{s} (\pi_{s}(p) $ from $S$ source images, \ie:
\begin{equation}
\centering
\begin{aligned} 
c=\sum_{s=1}^{S} w_{s}c_{s}.\\
\end{aligned}
\label{eq:blend}
\end{equation}

\paragraph{Volume rendering.}
For volume rendering along the rays, we follow the NeuS \cite{neus21} formulation to convert the SDF value $F^{s}(p)$ to the density $\sigma$ 
% for the accumulation of color on each ray
, \ie: 
\begin{equation}
\centering
\begin{aligned} 
\sigma=max(-\frac{ \frac{d\Phi }{dp} (\mathcal F^{s}(p) ) }{ \Phi (\mathcal F^{s }(p) )},0),
\end{aligned}
\label{eq:neus}
\end{equation}
where $p$ is the point sampled on the ray and $\Phi$ is the Sigmoid function.   
%This formulation improves the network generalization ability because it takes the spatial gradient of the current point SDF into consideration. Thus, we are also considering the SDF of the nearby points when predicting the density. Furthermore, we use Eikonal regularization in Eqn.~\ref{eq:eik} to regularize the SDF  space. As a result, our geometry prediction is spatial-aware because density learning considers the relations with the nearby points. 
Finally, for a ray sampled with M points, the color is accumulated as:
\begin{equation}
\centering
\begin{aligned} 
C(r)=\sum_{i=1}^{M}T_{i}(1-exp(-\sigma_{i}))_{i},\\
\text{where} ~~ T_{i}=exp(-\sum_{j=1}^{i-1} \sigma_{j}).
\end{aligned}
\label{eq:volume}
\end{equation}

\subsection{Score Distillation Sampling for Geometry Enhancement}
\begin{figure}[t] 
\centering 
\includegraphics[width=0.46\textwidth]{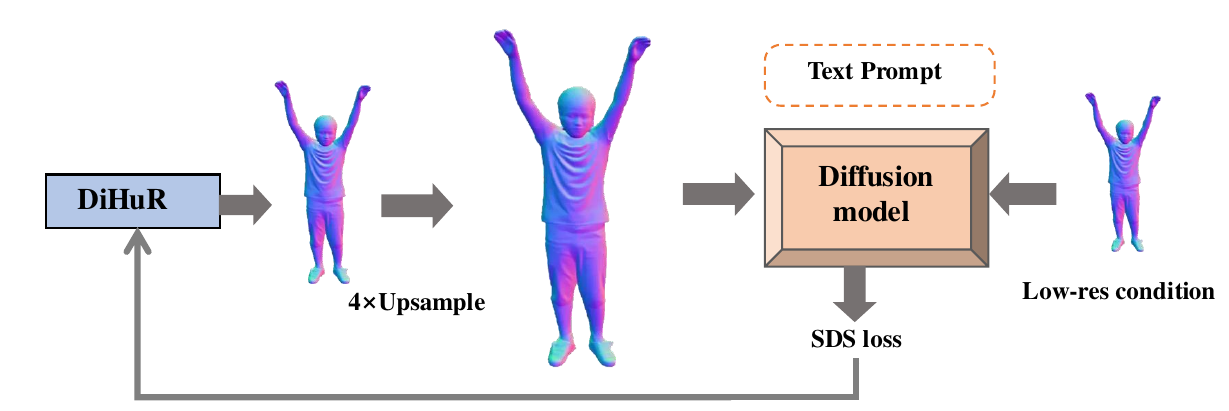} 
\vspace{-1mm}
\caption{We show how to compute SDS loss on the pre-trained super-resolution diffusion model: the low-resolution conditioned image on the right is generated from our fixed model.}
\vspace{-1mm}
\label{fig:sds}
\end{figure}

During the inference, as shown in Fig.~\ref{fig:sds}, we compute the $\mathcal{L}_{sds}$ \cite{dream23} with an pre-trained diffusion model. Different from the original SDS loss in \cite{dream23}, we choose to use a normal map as the input to the super-resolution diffusion model in order to enhance the geometric details. More specifically, we use our pretrained model to infer the normal map as the low-resolution image condition. 
We then fine-tune our model with the guidance from the diffusion model. Note that we only finetune $\mathcal{F}_{s}$ for fast convergence. We rendered the normal map and up-sample 4$\times$ for this normal map %, then we
and subsequently get the latent code from the VAE encoder in the diffusion model. The normal $N(r)$ for each ray is computed by volume rendering with the gradient of the points' SDF sampled on each ray as: 
% shown in Equ. \ref{eq:normal}, and the patch is selected with the center point randomly sampled on the body mask.
\begin{equation}
\centering
\begin{aligned} 
N(r)=\sum_{i=1}^{M}T_{i}(1-exp(-\sigma_{i})) \frac{\nabla \mathcal{S}(p_{i})}{||\nabla \mathcal{S}(p_{i})||},\\
\text{where} ~~ T_{i}=exp(-\sum_{j=1}^{i-1} \sigma_{j}),
\end{aligned}
\label{eq:normal}
\end{equation} 
We finally compute the SDS loss:
\begin{equation}
\centering
\begin{aligned} 
\mathcal{L}_{sds}=w(t)(\hat{\epsilon}_{\phi}(z_{t}  ;y,N^{low},t)-\epsilon)  \frac{\partial z_{t}}{\partial N} \frac{\partial N}{\partial \theta}, %z_{t}=Enc(N^{patch})
\end{aligned}
\label{eq:sds}
\end{equation}
with the input text and low-resolution normal map rendered from our fixed model as conditions.
$z_{t}$ is the noisy latent by adding noise to the latent from the up-sampled rendered normal map $N$, and $\theta$ are our model parameters. %nd $Enc$ is the pre-trained image encoder in the diffusion models. 
$\hat{\epsilon}_{\phi}$ is the noise prediction model and $\epsilon$ is the random noise added, t is sampled from 0.52 to 0.98 empirically, $N^{low}$ is the low-resolution normal map which is detached during the optimization. y is the text condition, we set it as `Best quality, human, normal map'. For classifier-free guidance(CFG), we set the guidance weights as 7.5. The intuition is to provide detailed normal supervision for our pre-trained network based on the 2D image super-resolution diffusion model. During SDS refinement, we select 8 views with azimuth evenly increasing from 0 to 360 degree to render the normal map. We also add image reconstruction loss for the input views as well as other regularization losses. The whole process takes 150 iterations around 2 minutes.

\subsection{Optimization}

\paragraph{Multi-target optimization.}
We believe multi-view images can serve as the surrogate for 3D supervision when direct supervision of the SDF  values is not available. Consequently, we use multiple images from different views as the target in each iteration instead of using one.
The random sampling strategy used in the original NeRF %might 
can lead to %non-overlap
non-intersecting rays between the multiple target views.
To solve this problem, we sample patches from the same body part in each target image to concentrate the optimization zone. Specifically, we simultaneously sample patches from the same segmentation part, \eg upper body, in the multiple target images at each iteration. This sampling strategy results in more overlapping points from different views, hence providing more explicit multi-view supervision. 
% we use 4 source images with minimal overlap FOV to train our model for each iteration. The training is done on pixels sampled from the same segmentation body part from 4-8 target view images.
The loss of our multi-target optimization is given by:
\begin{equation}
\centering
\begin{aligned} 
\mathcal{L}_{rgb}=\sum_{l=1}^{L}\sum_{r\in R} ||\hat C(r)_{l}-C(r)_{l} ||.
\end{aligned}
\label{eq:mul}
\end{equation}
$L$ is the number of target images for ray sampling and $R$ is the set of sampled rays for each target image.
\label{multar}

% \paragraph{Surface regularization.}
% To improve the geometry of the predicted surface, we add Eikonal regularization and a second-order SDF  regularization to guide the training process. Intuitively, the first-order gradient of SDF  represents the normal direction field of the space. We constraint the gradient strength to be 1 everywhere:

% For the second-order SDF  constraint, the surface tends to be smooth when we enforce the spatial gradient of the surface normal to be small. To achieve this, we enforce the normal vectors of nearby points to be similar with a smooth regularization, \ie :
% \begin{equation}
% \centering
% \begin{aligned} 
% \mathcal{L}_{sm}=\sum_{p \in S }|| \nabla \mathcal S(p) -\nabla \mathcal S(p+ \epsilon )   ||_{2}^{2},
% \end{aligned}
% \label{eq:sm}
% \end{equation}
% where $S$ represents the set of sampled points, and $\epsilon$ represents a perturbation %which is 
% sampled from a Gaussian distribution.

\paragraph{Total loss.}
We enforce the normal vectors of nearby points to be similar with a smooth regularization, \ie :
\begin{equation}
\centering
\begin{aligned} 
\mathcal{L}_{sm}=\sum_{p \in S }|| \nabla \mathcal S(p) -\nabla \mathcal S(p+ \epsilon )   ||_{2}^{2},
\end{aligned}
\label{eq:sm}
\end{equation}
where $S$ represents the set of sampled points, and $\epsilon$ represents a perturbation %which is 
sampled from a Gaussian distribution. We also include Eikonal regularization, \ie :
\begin{equation}
\centering
\begin{aligned} 
 \mathcal L_{eik}=  \sum_{p \in S}(||\nabla \mathcal F^{s }(p)  ||_{2}-1)^{2}. 
\end{aligned}
\label{eq:eik}
\end{equation} 
Our full objective function contains the color loss $\mathcal{L}_{rgb}$, SDS loss, the Eikonal regularization \cite{eik20} $\mathcal{L}_{eik}$, second-order SDF  regularization $\mathcal{L}_{sm}$
% and background loss with masks $\mathcal{L}_{bg}$ 
with corresponding weights as hyper-parameters, \ie:
\begin{equation}
\centering
\begin{aligned}
\mathcal{L}=\lambda_{rgb}\mathcal{L}_{rgb}+\lambda_{sds}\mathcal{L}_{sds}+\lambda_{eik}\mathcal{L}_{eik}+\lambda_{sm}\mathcal{L}_{sm}.
% \\+\lambda_{bg}\mathcal{L}_{bg}
\end{aligned}
\label{eq:loss}
\end{equation}
Note that only during the per-scene finetuning stage, we add SDS loss.
% \paragraph{Iterative sampling.}
% We use the SDF-based importance sampling \cite{neus21} with 8 initial points uniformly sampled between the near to the far bound. We then recurrently sampled 4 points with SDF  values close to 0 for $4\times$ based on the predicted SDF  values in the previous step.
% % We reduce the number of sampled points for each ray since we have more target views. In this way, the total number of sampling points is still less compared with the uniform sampling. 
% This helps accelerate the training and inference speed without losing the reconstruction accuracy.

\section{Experiments}
\subsection{Experiment Setup}
\paragraph{Dataset.}
We conduct experiments on the commonly used ZJU-MoCap dataset \cite{nb21}, THuman dataset \cite{thuman19} and HuMMan dataset\cite{ntuhuanman22}.
THUman contains 202 human
body 3D scans. Following \cite{gpnerf22}, $80\%$ of the scans are taken as the training set, and the remaining are the test set. For ZJU-MoCap dataset, it consists of 9 sequences captured with 23 calibrated cameras. Following \cite{gpnerf22}, we train our model using 6 sequences and test our model on the remaining 3 sequences for 3D reconstruction.
We further train our method on THuman dataset and conduct cross-dataset evaluation on HuMMan\cite{ntuhuanman22} dataset. Specifically, we use the last 22 sequences for evaluation.

% For Novel view synthesis, we follow \cite{gpnerf22} to use the same 6 sequences for training and the remaining 3 for testing. For THuman dataset, we follow MPS-NeRF \cite{mps22} taining 25 subjects and testing with 5 subjects.
% Note that masks provided in NB~\cite{nb21} are noisy, and thus background NeRF model can give more accurate results. 

\begin{figure*}
\centering
\includegraphics[width=0.9\textwidth]{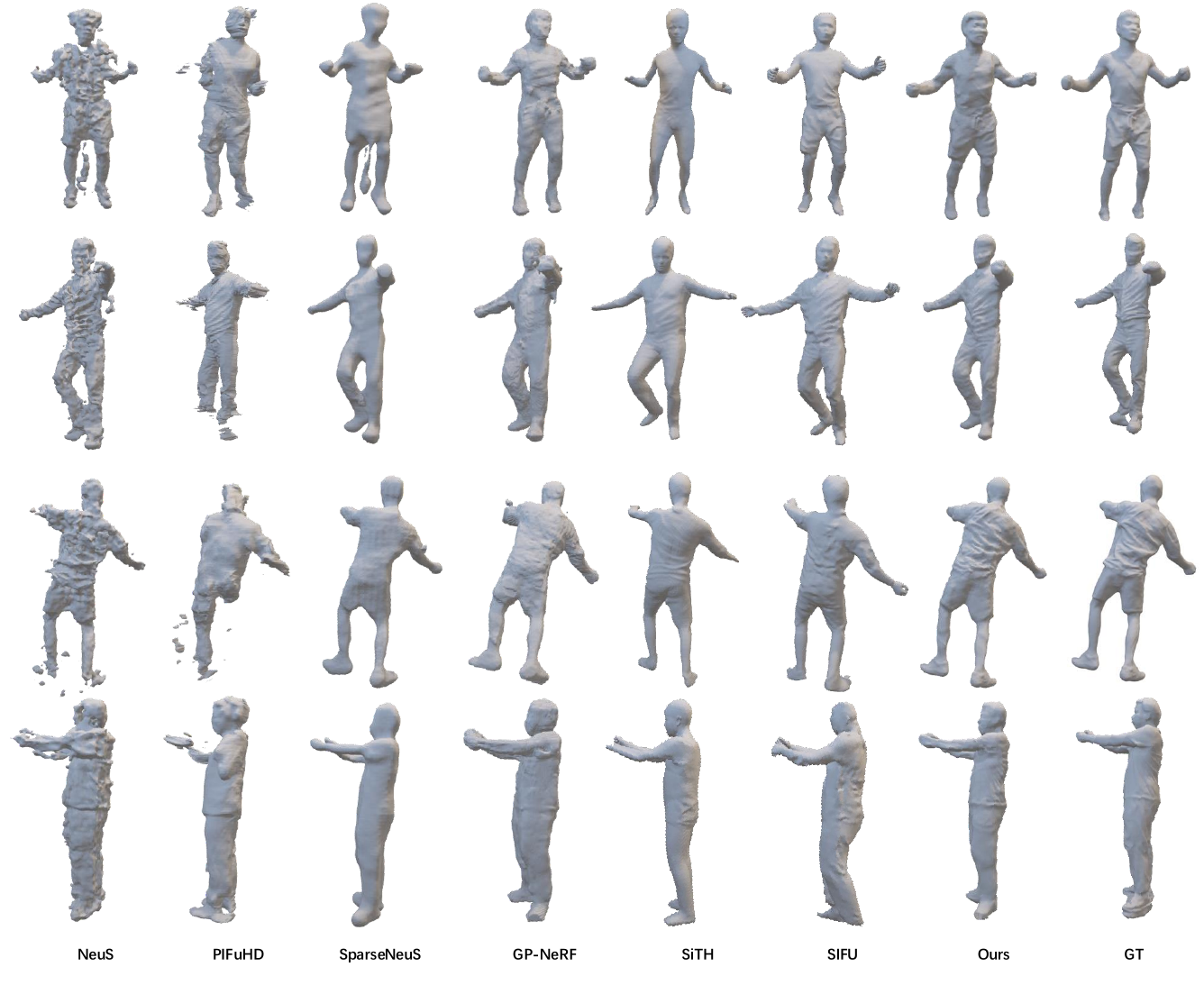} 
\vspace{-1mm}
\caption{
% Visual comparison with \cite{neus21,pifuhd20,sparseneus22,gpnerf22,sith24,sifu24}. 
From left to right are reconstruction results from NeuS\cite{neus21}, PIFuHD\cite{pifuhd20}, SparseNeuS\cite{sparseneus22}, GP-NeRF\cite{gpnerf22}, SiTH\cite{sith24}, SIFU\cite{sifu24}, Ours, and ground truth. For single view reconstruction methods\cite{pifuhd20,sith24,sifu24}, we choose the front view as input.}
\vspace{-1mm}
\label{fig:3dvis}
\end{figure*}

\begin{figure}
% \centering
\includegraphics[width=0.5\textwidth]{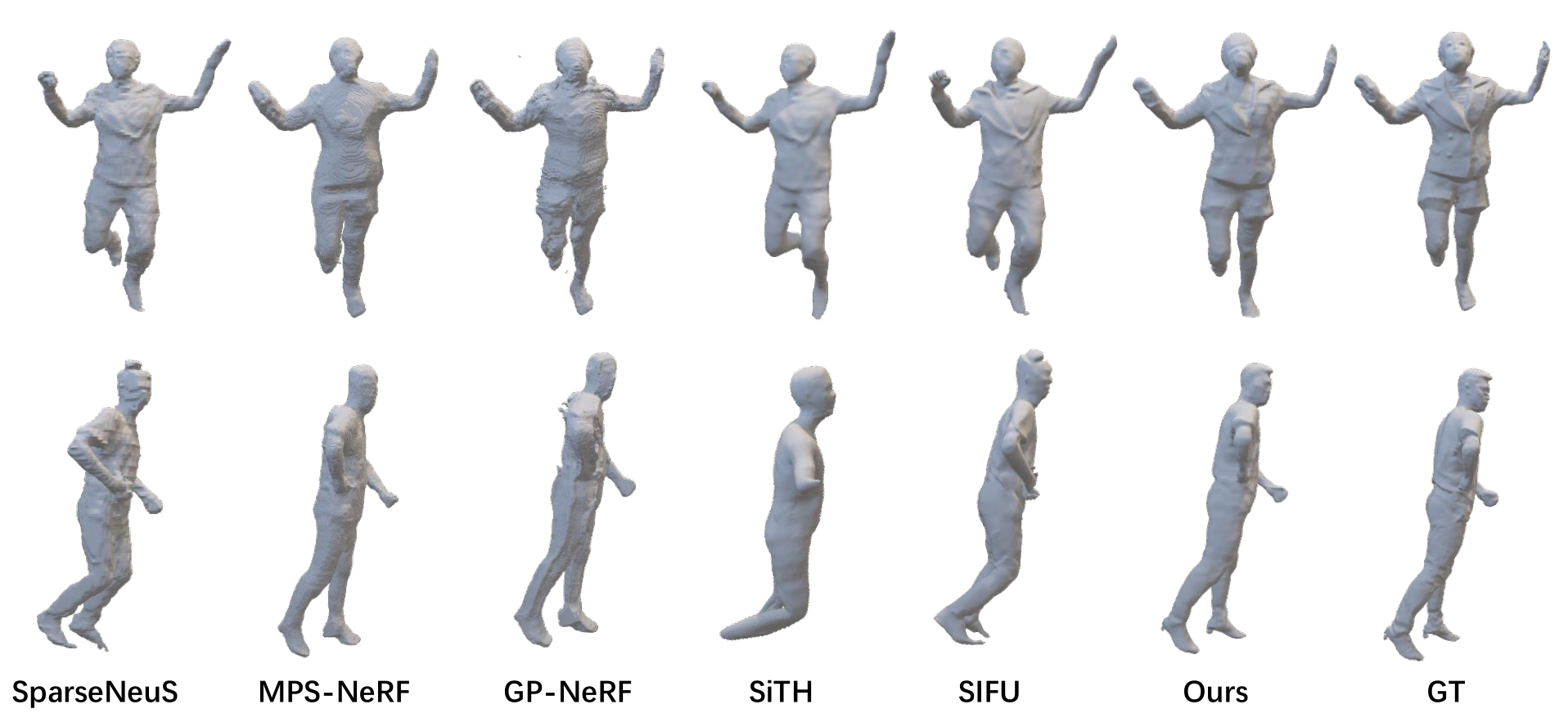} 
\caption{Visual comparison on THuman dataset. From left to right are results from SparseNeuS\cite{sparseneus22}, MPS-NeRF\cite{mps22}, GP-NeRF\cite{gpnerf22}, SiTH\cite{sith24}, SIFU\cite{sifu24}, Ours and ground truth.}
\label{fig:thu3dvis}
\end{figure}

\begin{figure}
% \centering
\includegraphics[width=0.45\textwidth]{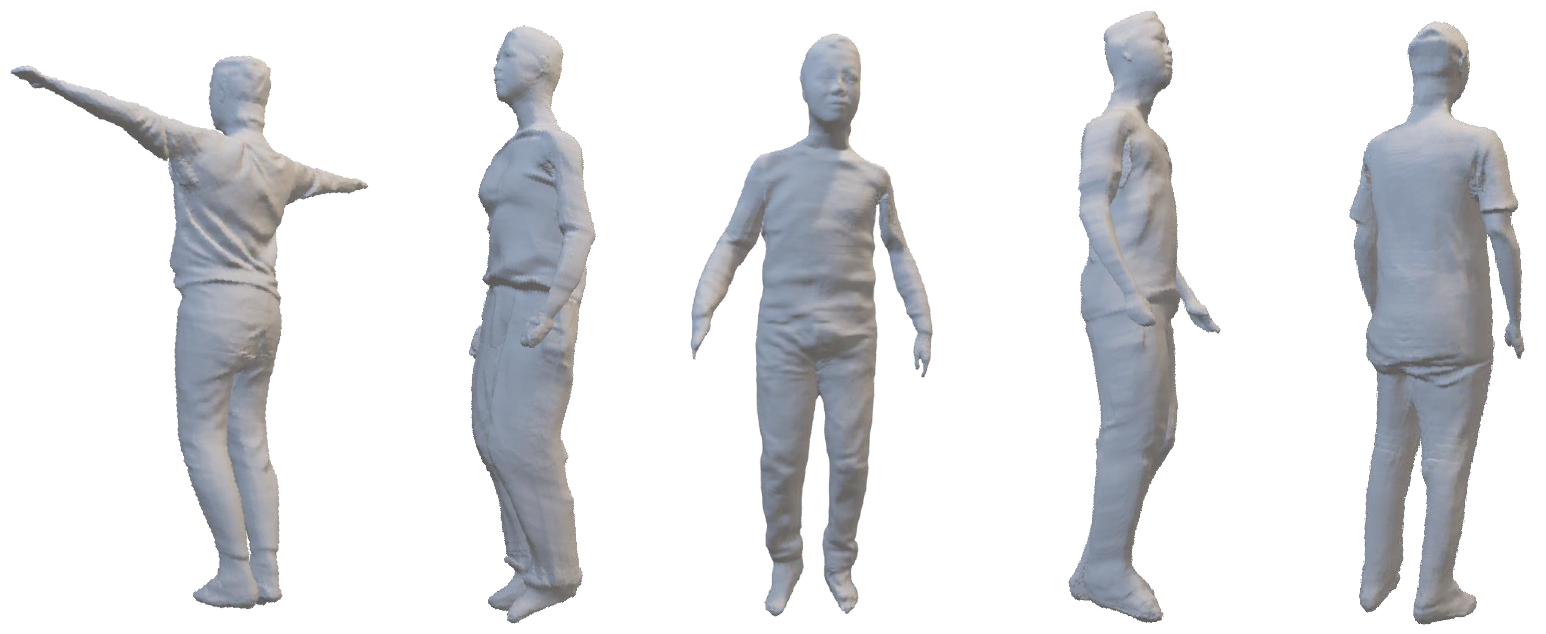} 
\caption{Zero-shot reconstruction results on HuMMan dataset.}
\label{fig:wild}
\end{figure}

\subsection{3D Reconstruction}
\paragraph{Baselines.}
Baselines: we compare our method with Generalizable NeRF methods: MPS-NeRF \cite{mps22}, SparseNeuS\cite{sparseneus22} and GP-NeRF \cite{gpnerf22} as well as single view human reconstruction methods PIFuHD\cite{pifuhd20}, SiTH\cite{sith24} and SIFU \cite{sifu24}. On ZJU-MoCap dataset, we also compared with NeuS\cite{neus21} and NB\cite{nb21}. Following \cite{gpnerf22}, we use 3 source views with minimal overlap FOV as the input. For single view reconstuction methods, we choose the front view as input. We use commonly used Chamfer Distance (CD) and Normal Consistency (NC) as the metrics for evaluation.
\paragraph{Evaluation.}
We show our 3D reconstruction results with other existing methods in Tab. \ref{table:thuman3D} for THuman dataset and Tab. \ref{table:3dcomp} for ZJU-MoCap. All the NeRF-based models are trained and tested on the same splits without 3D supervision.
% We train generalizable methods \cite{gpnerf22,sparseneus22,mps22} with the same data and directly do inference on both datasets.
Our methods outperform existing methods \cite{gpnerf22,sparseneus22,neus21,nb21,mps22,sifu24} with a large margin in terms of both CD and NC.
% which uses more information as the input and several hours of optimization. 
We show the qualitative comparison with existing methods in Fig.~\ref{fig:3dvis}. We can see in the figure that genralizable NeRF methods\cite{gpnerf22,mps22,sparseneus22} fail to reconstruct the details although the pose and shape are correct. Directly optimizing NeuS in this setting produces large noisy areas with disconnected points as shown in Fig. \ref{fig:3dvis}.  PIFuHD also shows poor generalization ability on this dataset with missing arms and legs as shown in Fig. \ref{fig:3dvis} with extremely large error, and therefore we do not include it in the quantitative results. SiTH and SIFU tend to reconstruct smooth meshes with wrong poses due to single view ambiguity, e.g. the second row in Fig. \ref{fig:3dvis} and Fig. \ref{fig:thu3dvis} compared with GT. In contrast, our model can infer a more realistic human mesh with fine details such as wrinkles on the clothes as well as correct body shape and pose. We also show cross-dataset results in Fig.\ref{fig:wild} to demonstrate the generalization ability, where our model is trained on ZJU-MoCap dataset.
\begin{table}[htbp]
\centering
\small
\setlength{\tabcolsep}{0.2cm}
\begin{tabular}{c|c|c|c|c|c|c|}
\hline
\multicolumn{1}{c|}{Methods}&{\cite{sparseneus22}}&{\cite{mps22}}&{\cite{gpnerf22}}&\cite{sith24}&\cite{sifu24} &{Ours}\\
\hline
% \multicolumn{1}{c|}{Components } & \multicolumn{1}{c|}{SSIM}&\multicolumn {1}{c|}{PSNR}&  \multicolumn {1}{c|}{CD}&{NC}\\
CD($\downarrow$)&  2.312&  3.312 &  3.876& 1.621& 1.521&  \textbf{1.117} \\
NC($\uparrow$)&  0.634&  0.616 &   0.567& 0.723 &0.741 & \textbf{0.779}  \\
\hline
\end{tabular} 

\caption{3D reconstruction comparison on THuman\cite{thuman19} dataset.}

\label{table:thuman3D}
\end{table}

\begin{table}[htbp]
\centering
\small
\setlength{\tabcolsep}{0.2cm}
\begin{tabular}{c|c|c|c|c|c|c|}
\hline
\multicolumn{1}{c|}{Methods}&{\cite{sparseneus22}}&{\cite{mps22}}&{\cite{gpnerf22}}&\cite{sith24}&\cite{sifu24} &{Ours}\\
\hline
% \multicolumn{1}{c|}{Components } & \multicolumn{1}{c|}{SSIM}&\multicolumn {1}{c|}{PSNR}&  \multicolumn {1}{c|}{CD}&{NC}\\
CD($\downarrow$)&  3.211&  3.412 &  3.765& 1.723& 1.709&  \textbf{1.23}  \\
NC($\uparrow$)&  0.633&  0.621 &   0.612& 0.711 &0.721 & \textbf{0.753}  \\
\hline
\end{tabular} 

\caption{Cross-dataset evaluation on HuMMan\cite{ntuhuanman22} dataset.}
\label{table:human3D}
\end{table}

\begin{table}[htbp]
\centering
\small
\setlength{\tabcolsep}{0.1cm}
\begin{tabular}{c|c|c|c|c|c|c|c}
% \toprule
\hline
 \multicolumn{1}{c|}{Methods}&\multicolumn {1}{c|}{\cite{nb21}}& \multicolumn {1}{c|}{\cite{neus21}}&  \multicolumn {1}{c|}{\cite{sparseneus22}}&  \multicolumn {1}{c|}{\cite{gpnerf22}}&  \multicolumn {1}{c|}{\cite{sith24}}&  \multicolumn {1}{c|}{\cite{sifu24}}& \multicolumn {1}{c}{Ours}\\

\hline

 CD($\downarrow)$& 1.876  &  4.544&  5.875 & 2.448&1.123&0.942&\textbf{0.790} \\
 ND($\uparrow)$& 0.624  &   0.478  &   0.403 &0.578 & 0.716 &0.707& \textbf{0.767}\\
 \hline
\end{tabular}
\caption{3D reconstruction comparison on ZJU-MoCap dataset.}

\label{table:3dcomp}
\end{table}
% For ZJU-MoCap dataset, we use the 3D reconstruction from NeuS with full 23 camera views as reference.
\begin{figure*}
\centering 
\includegraphics[width=1.0\textwidth]{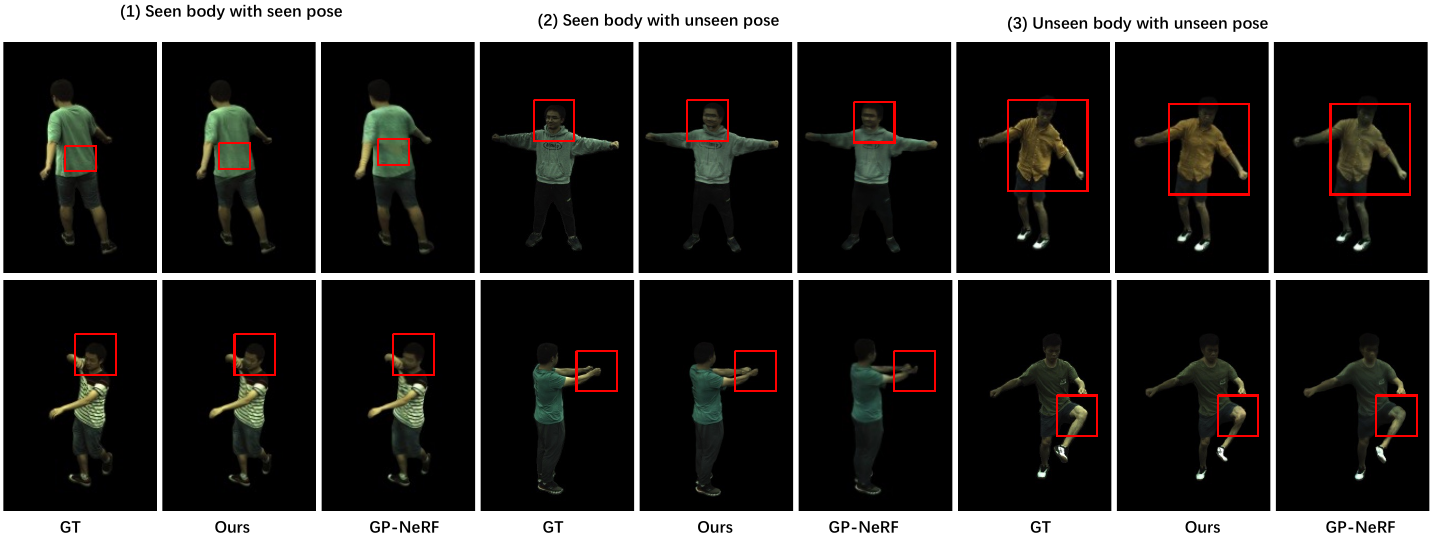} 
\vspace{-1mm}
\caption{Visual comparison with GP-NeRF \cite{gpnerf22}. From left to right are the results from: 1) Seen body with seen pose; 2) Seen body with unseen pose; 3) Unseen body with unseen pose. We show two examples for each setting. The details are highlighted in the red boxes.}
\vspace{-1mm}
\label{fig:2dvis}
\end{figure*}

\begin{table}[htbp]
\centering
\small
\setlength{\tabcolsep}{0.18cm}
\begin{tabular}{c|c|c|c|cc}
% \toprule
\hline
\multicolumn{1}{c|}{Method} &\multicolumn {1}{c|}{P.S.}&\multicolumn {1}{c|}{U.B.}&{U.P.}& \multicolumn {1}{c|}{PSNR($\uparrow$)}&{SSIM($\uparrow)$}\\
 % Method&Train &Test & Per-scene training &Unseen Pose Body& PSNR &SSIM\\
\hline
\multicolumn{6}{c}{Seen people on seen frames}\\
% NHR\cite{nhr20}&\cmark &\xmark &\xmark &23.95 &0.897\\
NB\cite{nb21}& \cmark &\xmark &\xmark &28.51 &0.947\\
MPS-NeRF\cite{mps22}&\xmark &\cmark &\cmark &28.54 &0.933\\
NHP\cite{nhp21}& \xmark&\xmark &\xmark  &28.73&0.936\\
GP-NeRF\cite{gpnerf22}&\xmark&\xmark &\xmark &28.81&0.944\\
Ours&\xmark &\xmark &\xmark & \textbf{28.94}&\textbf{0.951}\\
\hline
\multicolumn{6}{c}{Seen people on unseen frames}\\
NB\cite{nb21}&\cmark &\xmark &\cmark &23.79 &0.887\\
MPS-NeRF\cite{mps22}&\xmark &\cmark &\cmark &27.02 &0.931\\
NHP\cite{nhp21}&\xmark&\xmark &\cmark  &26.94&0.929\\
GP-NeRF\cite{gpnerf22}&\xmark&\xmark &\cmark &27.92&0.934\\
Ours&\xmark &\xmark &\cmark &\textbf{28.35} &\textbf{0.941}\\\\
\hline
\multicolumn{6}{c}{Unseen people on unseen frames}\\
NB\cite{nb21}&\cmark &\cmark &\xmark &22.88 &0.883\\
% PVA\cite{pva21}&\xmark &\cmark &\cmark &23.15 &0.866\\
MPS-NeRF\cite{mps22}&\xmark &\cmark &\cmark &25.17 &0.911\\
NHP\cite{nhp21}&\xmark&\cmark &\cmark  &24.75&0.906\\
GP-NeRF\cite{gpnerf22}&\xmark&\cmark &\cmark &25.96& 0.921\\
Ours &\xmark &\cmark &\cmark &\textbf{ 26.26}&\textbf{0.926}\\
\end{tabular} \vspace{2mm}
\caption{We thoroughly compared our method with existing NVS works under three settings on ZJU-MoCap dataset. P.S. refers to per-scene optimization, U.B. refers to unseen bodies, and U.P. refers to unseen poses.}
\label{table:2dcomp}
\end{table}

\subsection{Image Synthesis}
\paragraph{Baselines.}
For ZJU-MoCap, we show results of novel view synthesis in 3 source views with minimal overlap FOV setting and compare with existing NeRF-based methods \cite{nb21,nhp21,gpnerf22}. For THuman dataset, we follow \cite{mps22} and compare with existing human NeRF methods \cite{nb21,animatable21,gpnerf22,mps22}. We use commonly used SSIM and PSNR as evaluation metrics.
\begin{table}[htbp]
\centering
\small
\setlength{\tabcolsep}{0.11cm}
\begin{tabular}{|c|c|c|c|c|c|}
\hline
\multicolumn{1}{|c|}{Methods }&{NB\cite{nb21}}&{NHP\cite{nhp21}}&{MPS\cite{mps22}}&{GP-NeRF\cite{gpnerf22}}&{Ours}\\
\hline
% \multicolumn{1}{c|}{Components } & \multicolumn{1}{c|}{SSIM}&\multicolumn {1}{c|}{PSNR}&  \multicolumn {1}{c|}{CD}&{NC}\\
SSIM($\uparrow$)&  0.907 &  0.895& 0.914   & 0.923& \textbf{0.931}\\
PSNR($\uparrow$)&  24.86 &   24.10& 24.63  & 24.88  & \textbf{26.31} \\
\hline
\end{tabular} \vspace{1mm}
\caption{NVS comparison on THuman dataset.}
\label{table:thuman}
\end{table}
\paragraph{Evaluation.}
For ZJU-MoCap dataset, following GP-NeRF \cite{gpnerf22}, we show the generalization from three aspects: novel view synthesis on the seen bodies with seen poses, seen bodies with unseen poses, and unseen bodies. We show the results in Tab.~\ref{table:2dcomp} compared with the generalizable human novel view synthesis methods. As shown in the table, we achieve the best performance than all the other methods. For THuman dataset, we show the quantitative comparison with SOTA methods in Tab.~\ref{table:thuman}. We only test the unseen bodies on this dataset and achieve the best performance. We credit this to our more accurate 3D geometry modeling with SDS guidance and SMPL localized features.
We show the qualitative results in Fig.~\ref{fig:2dvis} only with \cite{gpnerf22}, which demonstrated the highest performance among all other methods in our evaluation in Tab.\ref{table:2dcomp}. Compared with GP-NeRF \cite{gpnerf22}, our method can synthesize images with better color and finer details. As highlighted in the red box of Fig.~\ref{fig:2dvis}, GP-NeRF generates wrong colors which are not consistent with the input images for some areas while the colors generated from our method show more consistency. While our method can synthesize images with more details. This is because with learned tokens as guidance, it can predict the accurate blending weights learned from the training data, which greatly reduces the blurry effect. Also, SDS optimization also helps color prediction since our blending weights prediction is relied on the accurate geometry prediction.
 % The qualitative comparisons on THuman dataset are in the supplementary.
\subsection{Ablation studies.}
We show the ablation studies for 3D reconstruction and novel view synthesis on THuman dataset in terms of learnable tokens, diffusion guidance, multi-target optimization in Tab.\ref{table:ablation}. As shown in the table, we can see that diffusion guidance plays an important role in both 3D reconstruction and novel view synthesis because our whole network is only supervised with images, and the whole network can converge better with the geometric SDS loss as guidance. %Then the learned network can have a stronger generalization ability to predict the unseen geometry during inference. 
Our learnable tokens also significantly improve both 3D and 2D results. Learnable tokens provides vertex-to-semantic mapping enables our model to generalize well to unseen subjects. Multi-target training makes the 3D geometry even better both quantitatively and qualitatively.
We also show the reconstruction results with different number of input views in Tab.\ref{table:ablation_views}. Increasing the number of input views, both NVS and 3D reconstruction quality are improved.
\begin{table}[htbp]
\centering
\small
\setlength{\tabcolsep}{0.11cm}
\begin{tabular}{c|c|c|c|c}
\hline
\multicolumn{1}{c|}{Number of views}&\multicolumn{2}{c|}{NVS}& \multicolumn{2}{c}{3D geometry}\\
\multicolumn{1}{c|}{ }&{PSNR($\uparrow$)}&{SSIM($\uparrow$)}&{CD($\downarrow$)}&{NC($\uparrow$})\\
\hline
% \multicolumn{1}{c|}{Components } & \multicolumn{1}{c|}{SSIM}&\multicolumn {1}{c|}{PSNR}&  \multicolumn {1}{c|}{CD}&{NC}\\
2&  24.98& 0.913&    1.345     & 0.702\\
3&  26.31 &    0.931    & 1.117 &0.779   \\
4&26.88 &0.938 &1.011 & 0.787 \\
5& \textbf{27.07}& \textbf{0.942} &\textbf{0.987} &\textbf{0.792} \\
\hline
\end{tabular} \vspace{1mm}
\caption{Ablation study for number of input views.}
\label{table:ablation_views}
\end{table}

\begin{figure}
\centering
\includegraphics[width=0.48\textwidth]{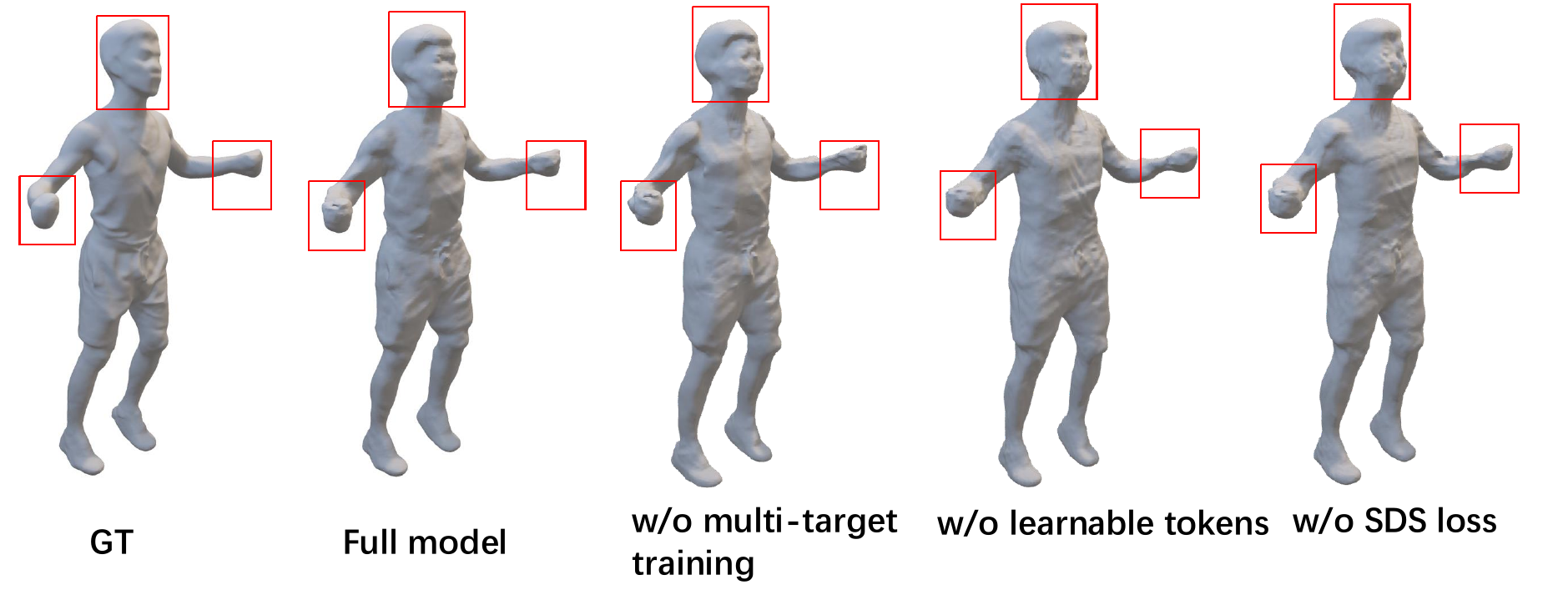} 
\vspace{-1mm}
\caption{Ablation studies on each component in our proposed method. See the detailed comparison highlighted in the red boxes.}
\vspace{-1mm}
\label{fig:ablation}
\end{figure}

\begin{table}[htbp]
\centering
\small
\setlength{\tabcolsep}{0.11cm}
\begin{tabular}{c|c|c|c|c}
\hline
\multicolumn{1}{c|}{Components}&\multicolumn{2}{c|}{NVS}& \multicolumn{2}{c}{3D geometry}\\
\multicolumn{1}{c|}{ }&{ PSNR($\uparrow$)}&{SSIM($\uparrow$)}&{CD($\downarrow$)}&{NC($\uparrow$})\\
\hline
% \multicolumn{1}{c|}{Components } & \multicolumn{1}{c|}{SSIM}&\multicolumn {1}{c|}{PSNR}&  \multicolumn {1}{c|}{CD}&{NC}\\
w/o leanable code&  23.98& 0.907&    1.543     & 0.667\\
w/o diffusion guidance&  25.58 &    0.921    & 1.344 &0.713   \\
w/o multi-target training&26.08 &0.926 & 1.243& 0.757     \\
full model& \textbf{26.31}& \textbf{0.931} &\textbf{1.117} &\textbf{0.779} \\
\hline
\end{tabular} \vspace{1mm}
\caption{Ablation study for each component.}
\label{table:ablation}
\end{table}
We also show the visual comparison in Fig.~\ref{fig:ablation} to validate the effectiveness of each component. We can see from the figure that without the SDS guidance, the 3D reconstruction suffers from losing details in both face and clothes, which is also the case for the results without learnable tokens as feature aggregation guidance. Multi-target optimization constraint compresses noisy part to make our reconstruction looks more natural. Details in the arms and facial part are highlighted in the red box. Our full model achieves precise 3D reconstruction, capturing fine details with high accuracy.

\section{Conclusion}
In this paper, we introduce DiHuR for generalizable 3D human reconstruction and novel view synthesis from sparse cameras. To mitigate the minimal overlapping view problem, we introduce learnable tokens attached on human parametric model to guide the sparse view feature aggregation and SDF prediction process. Without 3D supervision, we utilize a 2D pre-trained diffusion model as the normal guidance to improve geometry details. We further propose a multi-target training strategy to constrain the underlying 3D surface. Quantitative and qualitative results on the commonly used benchmark show the superiority of our DiHuR compared with existing approaches.
% In this paper, we propose to reconstruct the human mesh with sparse minimal overlap cameras without 3D supervision for more practical use. We utilize the SMPL model as the geometry prior to guiding the SDF  prediction, we propose to use a geometry code to guide the color blending process which also makes the 3D reconstruction better. We propose a multi-target training strategy with second-order SDF  regularization to constrain the underlying 3D surface. 

% In this paper, we propose to reconstruct the human mesh with sparse minimal overlap cameras without 3D supervision for more practical use. We utilize the SMPL model as the geometry prior to guiding the SDF  prediction, we propose to use a geometry code to guide the color blending process which also makes the 3D reconstruction better. We propose a multi-target training strategy with second-order SDF  regularization to constrain the underlying 3D surface. 
{\small
\bibliographystyle{ieee_fullname}
\bibliography{egbib}

\begin{thebibliography}{10}\itemsep=-1pt

\bibitem{ntuhuanman22}
Zhongang Cai, Daxuan Ren, Ailing Zeng, Zhengyu Lin, Tao Yu, Wenjia Wang, Xiangyu Fan, Yang Gao, Yifan Yu, Liang Pan, Fangzhou Hong, Mingyuan Zhang, Chen~Change Loy, Lei Yang, and Ziwei Liu.
\newblock {HuMMan}: Multi-modal 4d human dataset for versatile sensing and modeling.
\newblock In {\em ECCV}, 2022.

\bibitem{mvsnerf21}
Anpei Chen, Zexiang Xu, Fuqiang Zhao, Xiaoshuai Zhang, Fanbo Xiang, Jingyi Yu, and Hao Su.
\newblock Mvsnerf: Fast generalizable radiance field reconstruction from multi-view stereo.
\newblock In {\em CVPR}, 2021.

\bibitem{chen2024generalizable}
Jinnan Chen, Chen Li, Jianfeng Zhang, Lingting Zhu, Buzhen Huang, Hanlin Chen, and Gim~Hee Lee.
\newblock Generalizable human gaussians from single-view image.
\newblock {\em arXiv preprint arXiv:2406.06050}, 2024.

\bibitem{gpnerf22}
Mingfei Chen, Jianfeng Zhang, Xiangyu Xu, Lijuan Liu, Yujun Cai, Jiashi Feng, and Shuicheng Yan.
\newblock Geometry-guided progressive nerf for generalizable and efficient neural human rendering.
\newblock In {\em ECCV}, 2022.

\bibitem{mps22}
Xiangjun Gao, Jiaolong Yang, Jongyoo Kim, Sida Peng, Zicheng Liu, and Xin Tong.
\newblock Mps-nerf: Generalizable 3d human rendering from multiview images.
\newblock {\em PAMI}, 2022.

\bibitem{eik20}
Amos Gropp, Lior Yariv, Niv Haim, Matan Atzmon, and Yaron Lipman.
\newblock Implicit geometric regularization for learning shapes.
\newblock In {\em ICML}, 2020.

\bibitem{sith24}
Hsuan-I Ho, Jie Song, and Otmar Hilliges.
\newblock Sith: Single-view textured human reconstruction with image-conditioned diffusion.
\newblock In {\em CVPR}, 2024.

\bibitem{sherf23}
Shoukang Hu, Fangzhou Hong, Liang Pan, Haiyi Mei, Lei Yang, and Ziwei Liu.
\newblock Sherf: Generalizable human nerf from a single image.
\newblock {\em arXiv preprint:2303.12791}, 2023.

\bibitem{hu2023gauhumanarticulatedgaussiansplatting}
Shoukang Hu and Ziwei Liu.
\newblock Gauhuman: Articulated gaussian splatting from monocular human videos.
\newblock {\em arXiv preprint arXiv:2312.02973}, 2023.

\bibitem{tech24}
Yangyi Huang, Hongwei Yi, Yuliang Xiu, Tingting Liao, Jiaxiang Tang, Deng Cai, and Justus Thies.
\newblock {TeCH: Text-guided Reconstruction of Lifelike Clothed Humans}.
\newblock In {\em 3DV}, 2024.

\bibitem{nhp21}
Youngjoong Kwon, Dahun Kim, Duygu Ceylan, and Henry Fuchs.
\newblock Neural human performer: Learning generalizable radiance fields for human performance rendering.
\newblock {\em NeurIPS}, 2021.

\bibitem{magic23}
Chen-Hsuan Lin, Jun Gao, Luming Tang, Towaki Takikawa, Xiaohui Zeng, Xun Huang, Karsten Kreis, Sanja Fidler, Ming-Yu Liu, and Tsung-Yi Lin.
\newblock Magic3d: High-resolution text-to-3d content creation.
\newblock In {\em CVPR}, 2023.

\bibitem{liu2023humangaussian}
Xian Liu, Xiaohang Zhan, Jiaxiang Tang, Ying Shan, Gang Zeng, Dahua Lin, Xihui Liu, and Ziwei Liu.
\newblock Humangaussian: Text-driven 3d human generation with gaussian splatting.
\newblock {\em arXiv preprint arXiv:2311.17061}, 2023.

\bibitem{sparseneus22}
Xiaoxiao Long, Cheng Lin, Peng Wang, Taku Komura, and Wenping Wang.
\newblock Sparseneus: Fast generalizable neural surface reconstruction from sparse views.
\newblock In {\em ECCV}, 2022.

\bibitem{smpl15}
Matthew Loper, Naureen Mahmood, Javier Romero, Gerard Pons-Moll, and Michael~J. Black.
\newblock Smpl: a skinned multi-person linear model.
\newblock In {\em ICCV}, 2015.

\bibitem{nerf20}
Ben Mildenhall, Pratul~P. Srinivasan, Matthew Tancik, Jonathan~T. Barron, Ravi Ramamoorthi, and Ren Ng.
\newblock Nerf: Representing scenes as neural radiance fields for view synthesis.
\newblock In {\em ECCV}, 2020.

\bibitem{animatable21}
Sida Peng, Junting Dong, Qianqian Wang, Shangzhan Zhang, Qing Shuai, Xiaowei Zhou, and Hujun Bao.
\newblock Animatable neural radiance fields for modeling dynamic human bodies.
\newblock In {\em ICCV}, 2021.

\bibitem{nb21}
Sida Peng, Yuanqing Zhang, Yinghao Xu, Qianqian Wang, Qing Shuai, Hujun Bao, and Xiaowei Zhou.
\newblock Neural body: Implicit neural representations with structured latent codes for novel view synthesis of dynamic humans.
\newblock In {\em CVPR}, 2021.

\bibitem{dream23}
Ben Poole, Ajay Jain, Jonathan~T. Barron, and Ben Mildenhall.
\newblock Dreamfusion: Text-to-3d using 2d diffusion.
\newblock In {\em ICLR}, 2023.

\bibitem{diffusion22}
Robin Rombach, Andreas Blattmann, Dominik Lorenz, Patrick Esser, and Björn Ommer.
\newblock High-resolution image synthesis with latent diffusion models.
\newblock In {\em CVPR}, 2022.

\bibitem{pifu19}
Shunsuke Saito, Zeng Huang, Ryota Natsume, Shigeo Morishima, Angjoo Kanazawa, and Hao Li.
\newblock Pifu: Pixel-aligned implicit function for high-resolution clothed human digitization.
\newblock In {\em ICCV}, 2019.

\bibitem{pifuhd20}
Shunsuke Saito, Tomas Simon, Jason Saragih, and Hanbyul Joo.
\newblock Pifuhd: Multi-level pixel-aligned implicit function for high-resolution 3d human digitization.
\newblock In {\em CVPR}, 2020.

\bibitem{anerf21}
Shih-Yang Su, Frank Yu, Michael Zollh{\"o}fer, and Helge Rhodin.
\newblock A-nerf: Articulated neural radiance fields for learning human shape, appearance, and pose.
\newblock In {\em NeurIPS}, 2021.

\bibitem{neus21}
Peng Wang, Lingjie Liu, Yuan Liu, Christian Theobalt, Taku Komura, and Wenping Wang.
\newblock Neus: Learning neural implicit surfaces by volume rendering for multi-view reconstruction.
\newblock In {\em NeurIPS}, 2021.

\bibitem{ibrnet21}
Qianqian Wang, Zhicheng Wang, Kyle Genova, Pratul Srinivasan, Howard Zhou, Jonathan~T. Barron, Ricardo Martin-Brualla, Noah Snavely, and Thomas Funkhouser.
\newblock Ibrnet: Learning multi-view image-based rendering.
\newblock In {\em CVPR}, 2021.

\bibitem{arah22}
Shaofei Wang, Katja Schwarz, Andreas Geiger, and Siyu Tang.
\newblock Arah: Animatable volume rendering of articulated human sdfs.
\newblock In {\em ECCV}, 2022.

\bibitem{econ23}
Yuliang Xiu, Jinlong Yang, Xu Cao, Dimitrios Tzionas, and Michael~J. Black.
\newblock Econ: Explicit clothed humans optimized via normal integration.
\newblock In {\em CVPR}, 2023.

\bibitem{pixelnerf20}
Alex Yu, Vickie Ye, Matthew Tancik, and Angjoo Kanazawa.
\newblock {pixelNeRF}: Neural radiance fields from one or few images.
\newblock In {\em CVPR}, 2021.

\bibitem{sifu24}
Zechuan Zhang, Zongxin Yang, and Yi Yang.
\newblock Sifu: Side-view conditioned implicit function for real-world usable clothed human reconstruction.
\newblock In {\em CVPR}, 2024.

\bibitem{thuman19}
Zerong Zheng, Tao Yu, Yixuan Wei, Qionghai Dai, and Yebin Liu.
\newblock Deephuman: 3d human reconstruction from a single image.
\newblock In {\em ICCV}, 2019.

\end{thebibliography}
}
% \bibliography{egbib}
% \bibliography{egbib}
\end{document}